\documentclass{article}
\usepackage{spconf,amsmath,graphicx}
\usepackage{multirow}

\title{Cross-Modality Hashing with Partial Correspondence}
\name{Yun Gu, Haoyang Xue, Jie Yang}
\address{Institute of Image Processing and Pattern Recognition\\
		 Shanghai Jiao Tong Univerisity, Shanghai, China
		}

\begin{document}
\maketitle
\begin{abstract} 
	Learning a hashing function for cross-media search is very desirable due to its low storage cost and fast query speed. However, the data crawled from Internet cannot always guarantee good correspondence among different modalities which affects the learning for hashing function. In this paper, we focus on cross-modal hashing with partially corresponded data. The data without full correspondence are made in use to enhance the hashing performance. The experiments on Wiki and NUS-WIDE datasets demonstrates that the proposed method outperforms some state-of-the-art hashing approaches with fewer correspondence information. 
\end{abstract}
\begin{keywords}
	Cross-modality, Hashing, Partial Correspondence, Multimedia Search
\end{keywords}
\section{Introduction}\label{sec:intro}
	Hashing techniques are increasingly popular for scalable similarity search in many applications \cite{torralba2008small,rasiwasia2010new}. The basic task for hashing is to map the high-dimensional data into compact binary codes which is very effictive in Approximate Nearest Neighbour (ANN) search. A good hashing function should make the data objects similiar in orginal feature space have the same or similiar hash codes. 

	Many hashing approaches have been proposed in recent years. One of the most sucessful schemes is Local Sensitive Hashing (LSH) \cite{datar2004locality} which uses random projections to obtain the hashing functions. Inspired by the manifold theory, Sepctral Hashing \cite{weiss2009spectral} and its extensions attempt to capture the local manifold properties to learn hashing functions. However, these methods are designed for unimodal tasks which cannot seek the correlations between multiple modalities. 

	In modern multimedia retrieval, many applications involve data objects which consist of different modalities. For example, the wikipedia website provides both text and image descriptions for each entry. Learning the latent correlation between multiple modalities is valuable for cross-media retrieval. The task of image-to-image, image-to-text and text-to-image query can be combined into a unified framework. A lot of works focus on cross-modal hashing for fast similiarity research \cite{bronstein2010data,kumar2011learning,zhen2012co,zhen2012probabilistic}. Bronstein et.al \cite{bronstein2010data} firstly proposed the multimodal problem (MMSH) which learns the hashing function between the relevant modalities. Kumar et.al \cite{kumar2011learning} extends the spectral hashing to the multimodal scenery. Zhen et.al \cite{zhen2012co} proposed the CRH model which is learned by boosting algorithms. Zhen et.al \cite{zhen2012probabilistic} directly learns the binary codes with the latent variable models. 

\begin{figure*}
	\centering
	\includegraphics[width=0.7\textwidth]{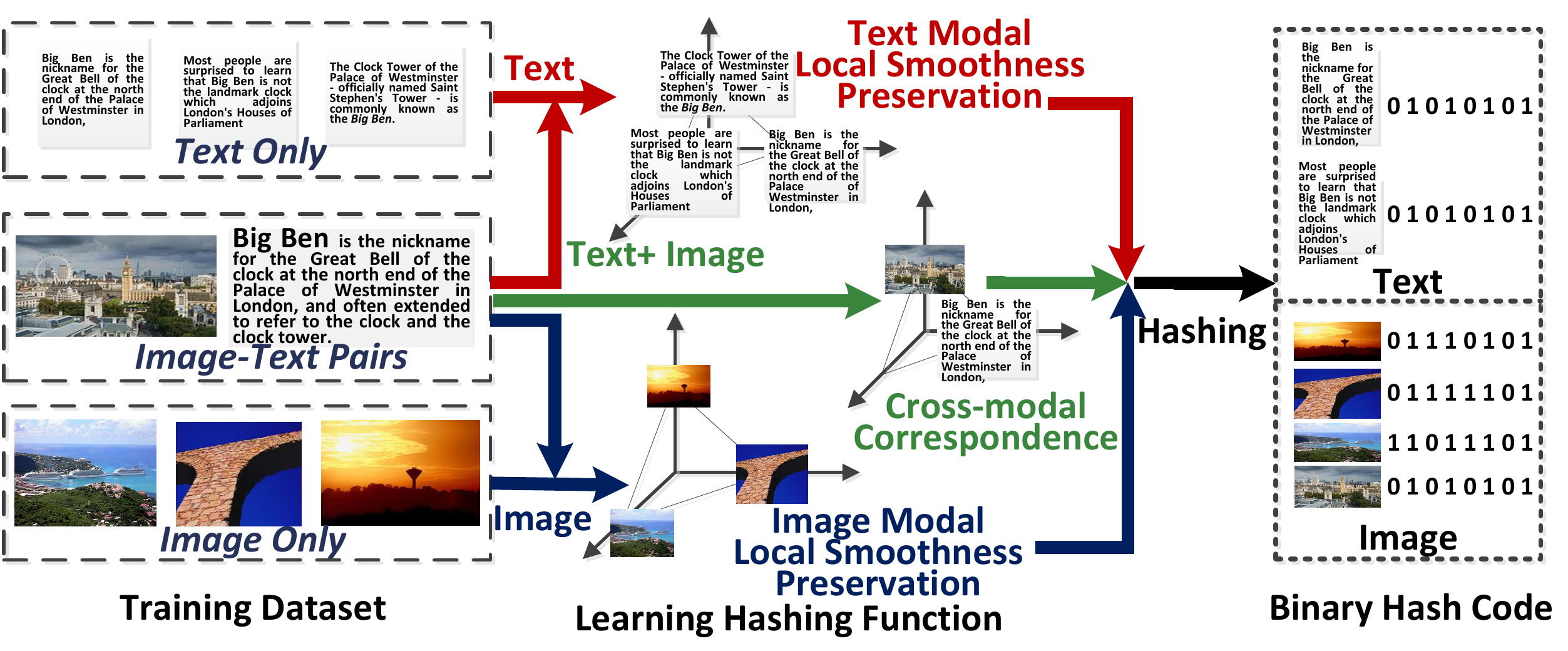}
	\caption{Main Framework of PCCMH. In each modality, an anchor graph is built to preserve the local smoothness. For well-corresponded data, we map the objects into Hamming spaces. The hashing function is learned via maximizing both the local smoothness and cross-modal correspondence.}\label{fig::mainframework}
\end{figure*}

	Although the multimodal \footnote{In this paper, cross-modal and multimodal refer to the same meaning.}hashing methods mentioned above have achieved good performance in many real datasets, they all require good correspondence between multiple modalities. More specifically, the image and text should be provided in pairs. However, a good matching between the text description and the image cannot be always guaranteed by the data crawled from Internet. The labels of images provided by user are usually with noisy information. The images in webpages might to be unavailable due to some transfering problems. It is very expensive to ensure that each image-text pair is in correspondence. In this paper, we focus on how to make use of the large amount of images and structural texts without fully correspondence in multiple modalities to enhance the hashing for cross-modal search. We propose Partial Correspondence Cross-Modal Hashing (PCCMH) to solve this problem as shown in Fig.\ref{fig::mainframework}. In each modality, an anchor graph \cite{liu2010large} is built to effectively capture the local manifold preserving the local smoothness. For well-corresponded data, we map the objects into Hamming spaces in which the modalities are tranformed into same or similiar binary codes. The hashing function is learned via maximizing the local smoothness in each modality and cross-modal correspondence.

	The remainder of this paper is organized as follows: The details of PCCMH is presented in Section \ref{sec::method}. In Section \ref{sec::exp}, we conduct extensive experiments to evaluate the performance of PCCMH in comparisons with some state-of-the-art methods. The conlusion and the discussion on future work are presented in Section \ref{sec::conclusion}. 

\section{Methodology}\label{sec::method}
	In this section, we describe the details of the proposed PCCMH method. We first focus on the definition of hashing problem with partially corresponded modalities in Section \ref{subsec::notation}. The details of local smoothness preservation and hashing with corresponded modalities are presented in Section \ref{subsec::non_corr} and \ref{subsec::corr}. Finally, we can derive the optimization problem of PCCMH in Section \ref{subsec::final_opt}.
\subsection{Notations and Problem Definition}\label{subsec::notation}
 	For ease of presentation, assume we have two modalities $M_x \subset \mathcal{R}^{d_x}$ and $M_y \subset \mathcal{R}^{d_y}$ where $d_x$ and $d_y$ are the dimensions of feature space in each modality. It is easy to extend our model to the hashing problem with more than two modalities. Let $X$ and $Y$ denote the training entities in two modalities where $X= \{x_1,x_2,...,x_n,x_{n+1},...,x_{n^{x}}\}, x_i \in M_x$ and $Y= \{y_1,y_2,...,y_n,y_{n+1},...,y_{n^{y}}\}, y_i \in M_y$. A fraction of data in $X$ and $Y$ are corresponded pairs indicating the same objects which are represented by $\{x_i,y_i\},i=1,2,...,n$ while the rest of them are not. The task for PCCMH is to learn two hashing functions for two modalities: $f(x): \mathcal{R}^{d_x} \to \{-1,1\}^c$ and $g(y): \mathcal{R}^{d_y} \to \{-1,1\}^c$, where $c$ is the length of the binary hash code\footnote{In order to obtain a feasible solution, we use $\{-1,1\}$ instead of $\{0,1\}$ as binary code.}. These two functions map the feature vectors in each modality into the same Hamming space. The goal is to make the data objects similiar in multiple modalities have the same or similiar hash codes.

\subsection{Local Smoothness Preserving with Non-corresponded Modality}\label{subsec::non_corr}
	As shown in Fig.\ref{fig::mainframework}, we tend to capture the local smoothness as Spectral Hashing \cite{weiss2009spectral} does which can effective make use of the partially and fully corresponded data. The data similiar in original feature space should be similiar in Hamming space after hashing. The first step is to build the smiliarity matrix $W$ in each modality. However, this procedure takes $O(n^2)$ time complexity where $n$ is the number of data. It is unacceptable for large-scale dataset. Therefore, we adopt the Anchor Graph \cite{liu2010large} to exploit the similiarity of the data effectively with $O(n)$ complexity.

	Take the modality $X$ as example, the anchor graph uses $m_x (m_x\ll n_x)$ landmarks (i.e. anchors)$\{u_i\}, i=1,...,m, u_i \in X$. The anchors can be effectively obtained by scalable K-mean algorithm \cite{chen2011large} with $O(n_x)$ complexity. The similarity between the training data and anchors can be computed as:
	\begin{equation}
		Z_{x}(i,j)=e^{-\|x_i-u_j\|^2/\sigma},\quad i=1,...,n_x, j=1,...,m
	\end{equation}
	Therefore, each data point is transformed into a new vector whose element is the similiarity between the data and the anchors.

	With the support from anchor graph, the similarity matrix $W_x$ can be approximately obtained as follows:
	\begin{equation} \label{eq::anchor}
	\widetilde{W_x}=Z_x\Lambda^{-1}(Z_x)^T, \Lambda = diag((Z_x)^T\mathbf{1}) 
	\end{equation}

	In order to preserve the local smoothness in hashing, we minimize the following equation:
	\begin{equation}\label{eq::smooth}
		 \min \sum_{i,j=1}^{n_x}W^{x}(i,j)\|f(x_i)-f(x_j)\|^2
	\end{equation}

	Although many kinds of functions can be used to define $f(x)$, we adopt the commonly-used linear hashing scheme: $f(x)=sgn(xB_{x})$ where $sgn(\cdot)$ denotes the element-wise sign function and $B_x$ is a linear matrix to learn. With the help from Anchor Graph, we have $f(X)=sgn(XB_{x})=sgn(Z_x\widetilde{B_x}) ,X=\{x_1,...,x_{n_x}\}$. According to Eq. \ref{eq::anchor}, the local smoothness in Eq. \ref{eq::smooth} can be rewritten as follows in matrix form:
	\begin{equation}\label{eq::smooth2}
		\min \quad tr(sgn(Z_x\widetilde{B_x})^TL_xsgn(Z_x\widetilde{B_x}))
	\end{equation}
	where $L_x=D_x-W_x$ and $D_x$ is the diagnal matrix whose elements are sum of each row in $W_x$. It is NP-hard to directly obtain the hashing problem in Eq. \ref{eq::smooth2}. We apply the relaxing form \cite{weiss2009spectral} to solve it:
	\begin{eqnarray}\label{eq::smooth3}
		\begin{aligned}
		\min & \quad tr((Z_x\widetilde{B_x})^TL_x(Z_x\widetilde{B_x}))\\
		s.t. & \quad (\widetilde{B_x})^T(\widetilde{B_x})=n_x\mathbf{I}_c
		\end{aligned}
	\end{eqnarray}
	where $I_c$ denotes an identity matrix of size $c\times c$.
	
	Since $W_x$ can be approximated with $Z$, Eq. \ref{eq::smooth3} can be written as:
	\begin{eqnarray}\label{eq::smooth4}
		\begin{aligned}
		\min & \quad tr((\widetilde{B_x})^T\widetilde{L_x}(\widetilde{B_x}))\\
		s.t. & \quad (\widetilde{B_x})^T(\widetilde{B_x})=n_x\mathbf{I}_c
		\end{aligned}
	\end{eqnarray}
	where $\widetilde{L_x}=(Z_x)^T(Z_x)-(Z_x)^T(Z_x)\Lambda^{-1}(Z_x)^T(Z_x)$ is the reduced graph Laplacian. In order to maximize the local smoothness in both two modalities, we have:
	\begin{eqnarray}\label{eq::smooth_final}
		\begin{aligned}
		\min & \quad tr((\widetilde{B_x})^T\widetilde{L_x}(\widetilde{B_x}))+tr((\widetilde{B_y})^T\widetilde{L_y}(\widetilde{B_y}))\\
		s.t. & \quad (\widetilde{B_x})^T(\widetilde{B_x})=n_x\mathbf{I}_c\\
			 & \quad (\widetilde{B_y})^T(\widetilde{B_y})=n_y\mathbf{I}_c
		\end{aligned}
	\end{eqnarray}
	where $\widetilde{L_y}$ is the reduced graph Laplacian obtained by the Anchor Graph in modality $Y$.

\subsection{Hashing with Corresponded Modalities}\label{subsec::corr}
	For the training data in good correspondence (i.e. data pairs $\{x_i,y_i\},i=1,2,...,n$ are provided), we firstly transform them with the Anchor Graph into similarity matrix $Z_x^m \in \mathcal{R}^{n\times m_x}$ and $Z_y^m \in \mathcal{R}^{n\times m_y}$ where $m_x$ and $m_y$ are the numbers of anchors in two modalities respectively. For the data pair $\{x_i,y_i\}$, the hash code should be similiar in Hamming space. The maximization of the cross-modal correspondence can be obtained via the following optimization problem:
	\begin{equation}\label{eq::binary1}
		\min \quad \|sgn(Z_x^m\widetilde{B_x})-sgn(Z_y^m\widetilde{B_y})\|^2_F
	\end{equation}

	Using the relaxing form mentioned in Section \ref{subsec::non_corr}, th optimization problem is rewritten as follows:
	\begin{eqnarray}\label{eq::binary_final}
		\begin{aligned}
		\min & \quad \|Z_x^m\widetilde{B_x}-Z_y^m\widetilde{B_y}\|^2_F\\
		s.t. & \quad (\widetilde{B_x})^T(\widetilde{B_x})=n_x\mathbf{I}_c\\
			 & \quad (\widetilde{B_y})^T(\widetilde{B_y})=n_y\mathbf{I}_c
		\end{aligned}
	\end{eqnarray}
	The optimization problem above can minimize the difference on two representation of an object in different modality.

\subsection{Final Optimization Problem}\label{subsec::final_opt}
	Combining optimization problems in Eq.\ref{eq::smooth_final} and \ref{eq::binary_final}, we have the final problem as follows:
	\begin{eqnarray}\label{eq::final1}
		\begin{aligned}
		\min & \quad \|Z_x^m\widetilde{B_x}-Z_y^m\widetilde{B_y}\|^2_F\\
			 &+\lambda(tr((\widetilde{B_x})^T\widetilde{L_x}(\widetilde{B_x}))+tr((\widetilde{B_y})^T\widetilde{L_y}(\widetilde{B_y})))\\
		s.t. & \quad (\widetilde{B_x})^T(\widetilde{B_x})=n_x\mathbf{I}_c\\
			 & \quad (\widetilde{B_y})^T(\widetilde{B_y})=n_y\mathbf{I}_c
		\end{aligned}
	\end{eqnarray}
	where $\lambda$ is the balance coefficient between local smoothness and cross-modality hashing. We can transform the cross-modality term as follows:
	\begin{eqnarray}\label{eq::final2}
		\begin{aligned}
		 & \|Z_x^m\widetilde{B_x}-Z_y^m\widetilde{B_y}\|^2_F\\
		= & tr((Z_x^m\widetilde{B_x}-Z_y^m\widetilde{B_y})^T(Z_x^m\widetilde{B_x}-Z_y^m\widetilde{B_y}))
		\end{aligned}
	\end{eqnarray}
	Therefore, the final objective function is:
	\begin{eqnarray}\label{eq::final3}
		\begin{aligned}
		 & \quad \|Z_x^m\widetilde{B_x}-Z_y^m\widetilde{B_y}\|^2_F\\
			 &+\lambda(tr((\widetilde{B_x})^T\widetilde{L_x}(\widetilde{B_x}))+tr((\widetilde{B_y})^T\widetilde{L_y}(\widetilde{B_y})))\\
		= & tr((Z_x^m\widetilde{B_x}-Z_y^m\widetilde{B_y})^T(Z_x^m\widetilde{B_x}-Z_y^m\widetilde{B_y})\\
		  &	+\lambda(tr((\widetilde{B_x})^T\widetilde{L_x}(\widetilde{B_x}))+tr((\widetilde{B_y})^T\widetilde{L_y}(\widetilde{B_y})))\\
		= & tr(B^TZB+\lambda B^TLB)\\
		\end{aligned}
	\end{eqnarray}
	where $B=[\widetilde{B_x}^T; \widetilde{B_y}^T]^T$ and 
	\begin{displaymath}
		\begin{array}{ll}
			Z=
			\left [
				\begin{array}{ll}
					(Z_x^m)^T(Z_x^m) & -(Z_x^m)^T(Z_y^m) \\
					-(Z_y^m)^T(Z_x^m) & (Z_y^m)^T(Z_y^m)
				\end{array}
			\right ] &
			L=
			\left [
				\begin{array}{ll}
					\widetilde{L}_{x} & \mathbf{0} \\
					\mathbf{0} & \widetilde{L}_{y}
				\end{array}
			\right ]
		\end{array}
	\end{displaymath}

	The final optimization problem in Eq.\ref{eq::final1} becomes:
	\begin{eqnarray}\label{eq::final}
		\begin{aligned}
		\min & \quad tr(B^T(Z+\lambda L)B)\\
		s.t. & \quad B^TB=\mathbf{I}
		\end{aligned}
	\end{eqnarray}

	Eq.\ref{eq::final} is an eigenvalue problem. The optimal $W$ is the eigenvectors corresponding to the $2c$ smallest eigenvalues of $Z+\lambda L$. We can solve it to obtain the hashing functions $f(x)$ and $g(y)$. 

\section{Experiments}\label{sec::exp}
	In this section, we evalute the performance of PCCMH by conducting extensive experiments and compare it with some state-of-the-art methods. All experiments are conducted on a workstation with Xeon (R) CPU W3503@2.40GHz and 6GB RAM. 

\subsection{Datasets}
	We conduct the experiments on two widely-used datasets:
	\begin{itemize}
		\item \textbf{Wiki}: Wiki-dataset is generated from 2866 Wikipedia documents \cite{rasiwasia2010new}. Each document is a couple of image and text keywords. The images are represented by 128D SIFT BoW features while the text key words are represented by 10D features whose elements are topics extracted by Latent Dirichlet Allocation.
		\item \textbf{NUS-WIDE}: NUS-WIDE dataset is a large-scale web image dataset crawled from Flickr \cite{chua2009nus}. It contains 269,648 images associated with 81 semantic concepts. Each image is represented by 500D SIFT BoW features and 1000D textual feature obtained by performing PCA on the original tag occurrence features.
	\end{itemize}

\subsection{Baselines and Evaluation Scheme}
	In this paper, we evaluate the proposed method with two typical cross-modal retrieval tasks: querying image database
	by some text words (text-to-image) and query the text database by images (image-to-text). Some state-of-the-art cross-modal hashing schemes are used for comparison	including MMSH\cite{bronstein2010data}, CCA\cite{gong2011iterative}, CRH\cite{zhen2012co} and MLBE\cite{zhen2012probabilistic}.

	The performance of retrieval is evaluated with Mean Average Precision (MAP).For a query $q$,the average precision is defined as:
	\begin{equation}
		AP(q)=\frac{1}{L_q}\sum_{r=1}^{R}P_q(r)\delta_q(r)
	\end{equation}
	where $L_q$ is the number of ground truth neighbours in the retrieved list. $P_q(r)$ is the precision of the top $r$ retrieved results and $\delta_q(r)=1$ if the $r$-th result is the true neighbour and 0 otherwise. $R$ is set to 50 in this paper.
		\begin{figure}[!t]
			\includegraphics[width=0.5\textwidth]{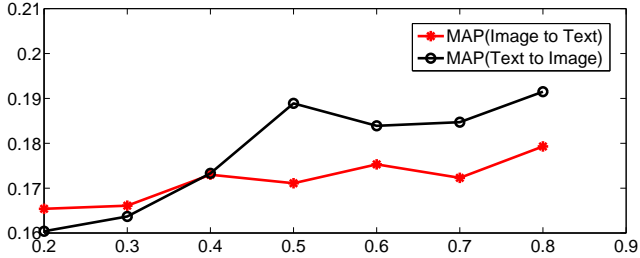}
			\caption{MAP Result of PCCMH on Wiki Dataset with different amount of corresponding information}\label{fig::map}
		\end{figure}

\subsection{Accuracy on Wiki Dataset}
	For Wiki Dataset, we conduct two experiments for evaluation: (1) We randomly select 60\% of the training data of which the correspondence in both text and image modality is available. For the rest, the correspondence is not provided for training. (2) We vary the number of corresponded data from 20\% to 80\% of the whole dataset to evaluate the performance of PCCMH. For other baselines, the whole correspondences are given. The number of landmarks ($m_x$ and $m_y$) in anchor graph in 200. The balance coefficient in Eq.\ref{eq::final} is 0.6. The experiment is repeated for 10 times. As shown in Table \ref{tab::wiki}, PCCMH achieves the best or closed performance in comparison with the baselines when 60 \% of correspondence information is available. 
	\begin{table}[!ht]
		\centering
		 \caption{MAP Result on Wiki Dataset}\label{tab::wiki}
		\begin{tabular}{|c|c|c|c|c|}
		 \hline
		 \multirow{2}{*}{Task} &
		 \multirow{2}{*}{Method} &
		 \multicolumn{3}{c|}{Code Length (bit)} \\
		 \cline{3-5}
		 &  & $c=16$ & $c=24$ & $c=32$ \\
		 \hline
		 \hline
		 \multirow{4}{2.2cm}{ \\ Image Query  \\ v.s. \\ Text Dataset}  
                & PCCMH & \textbf{0.1753}& \textbf{0.1774} & 0.1657 \\
                \cline{2-5}
                & CCA   & 0.1658         & 0.1532 & 0.1558 \\
                \cline{2-5}
                & CRH   & 0.1370         & 0.1605 & 0.1398 \\
                \cline{2-5}
                & MLBE  & 0.1573         & 0.1751 & \textbf{0.1793} \\
                \cline{2-5}
                & MMSH  & 0.1684         & 0.1617 & 0.1624 \\
		 \hline
		 \hline
		 \multirow{4}{2.2cm}{ \\ Text Query  \\ v.s. \\ Image Dataset}  
                & PCCMH & \textbf{0.1839} & \textbf{0.2010} & 0.1904 \\
                \cline{2-5}
                & CCA   & 0.1658 & 0.1532& 0.1558\\
                \cline{2-5}
                & CRH   & 0.1341 & 0.1605& 0.1398 \\
                \cline{2-5}
                & MLBE  & 0.1827 & 0.1624& \textbf{0.2107} \\
                \cline{2-5}
                & MMSH  & 0.1707 & 0.1824& 0.1724 \\
		 \hline
		 \end{tabular}
	\end{table}
	
	The performance of PCCMH with varying correspondence information is presented in Fig.\ref{fig::map}. When varying the ratio of corresponded data from 20\% to 80\%, the performance of PCCMH is increasing both in text-to-image and image-to-text retrieval. However, the improvement is not obvious with the ratio larger than 50\%. Since the number of class in Wiki dataset is limited, 50 \% of the dataset can cover most of diversity. The preservation of local smoothness provides good generalization of the hashing model. 

\subsection{Accuracy on NUS-WIDE Dataset}
	We randomly select 5,000 samples from NUS-WIDE dataset in which 4,000 samples are used for training and the rest are used for testing. For PCCMH, 60\% of correspondence information is available. The parameters of PCCMH for NUS-WIDE is as same as Wiki dataset. As shown in Table \ref{tab::nus}, PCCMH still outperforms the baselines in most cases with 60\% of correspondence information. 
	\begin{table}[!ht]
		\centering
		 \caption{MAP Result on NUS-WIDE Dataset}\label{tab::nus}
		\begin{tabular}{|c|c|c|c|c|}
		 \hline
		 \multirow{2}{*}{Task} &
		 \multirow{2}{*}{Method} &
		 \multicolumn{3}{c|}{Code Length (bit)} \\
		 \cline{3-5}
		 &  & $c=16$ & $c=24$ & $c=32$ \\
		 \hline
		 \hline
		 \multirow{4}{2.2cm}{ \\ Image Query  \\ v.s. \\ Text Dataset}  
                & PCCMH & \textbf{0.2334} & \textbf{0.2281} & 0.2172 \\
                \cline{2-5}
                & CCA   & 0.2214 & 0.2212 & 0.2264 \\
                \cline{2-5}
                & CRH   & 0.2123 & 0.2010 & 0.1865 \\
                \cline{2-5}
                & MLBE  & 0.1840 & 0.2030 & 0.2222 \\
                \cline{2-5}
                & MMSH  & 0.1920 & 0.2245& \textbf{0.2364} \\
		 \hline
		 \hline
		 \multirow{4}{2.2cm}{ \\ Text Query  \\ v.s. \\ Image Dataset}  
                & PCCMH & \textbf{0.2261} & 0.2311& \textbf{0.2349} \\
                \cline{2-5}
                & CCA   & 0.2208 & 0.2211& 0.2285\\
                \cline{2-5}
                & CRH   & 0.2233 & 0.2305& 0.2260 \\
                \cline{2-5}
                & MLBE  & 0.2188 & 0.2064& 0.2226 \\
                \cline{2-5}
                & MMSH  & 0.1920 & \textbf{0.2425}& 0.2164 \\
		 \hline
		 \end{tabular}
	\end{table}
\section{Conclusion}\label{sec::conclusion}
	Most existing cross-modal hashing methods need full correspondended data to learn the hashing function. In this paper, we propose PCCMH using partially correponded information for cross-modal hashing. Experiments on Wiki and NUS-WIDE dataset demonstrate the feasibility and good performance of PCCMH.


\bibliographystyle{IEEEbib}
\bibliography{refs}

\end{document}